\documentclass[11pt,a4paper]{article}
\usepackage[hyperref]{emnlp-ijcnlp-2019}
\usepackage{times}
\usepackage{latexsym}
\usepackage{graphicx}

\usepackage{url}

\aclfinalcopy 

\let\OLDthebibliography\thebibliography
\renewcommand\thebibliography[1]{
  \OLDthebibliography{#1}
  \setlength{\parskip}{0pt}
  \setlength{\itemsep}{1ex plus 1ex}
}

\title{Named Entity Recognition - Is there a glass ceiling?}

\author{Tomasz Stanislawek$^{\dagger,\ddagger}$, Anna Wr\'oblewska$^{\dagger,\ddagger}$, Alicja W\'ojcicka$^{\dagger,\mathsection}$, \\\textbf{Daniel Ziembicki$^{\dagger,\mathsection}$ Przemyslaw Biecek}$^{\ddagger,\mathparagraph}$\\ \\
  $^\dagger$Applica.ai, Warsaw, Poland\\ 
   $^\mathparagraph$Samsung R\&D Institute Poland, Warsaw, Poland\\ 
   $^\ddagger$Faculty of Mathematics and Information Science,
   Warsaw University of Technology\\ 
   $^\mathsection$Department of Formal Linguistics, University of Warsaw
}

\date{}

\begin{document}
\maketitle
\begin{abstract}
  Recent developments in Named Entity Recognition (NER) 
have resulted in better and better models. However, is there a glass ceiling? Do we know which types of errors are still hard or even impossible to correct? In this paper, we present a detailed analysis of the types of errors in state-of-the-art machine learning (ML) methods. Our study reveals the weak and strong points of the Stanford, CMU, FLAIR, ELMO and BERT models, as well as their shared limitations. We also introduce new techniques for improving annotation, for training processes and for checking a model's quality and stability.

Presented results are based on the CoNLL 2003 data set for the English language. A new enriched semantic annotation of errors for this data set and new diagnostic data sets are attached in the supplementary materials. 
\end{abstract}

\section{Introduction}

The problem of Named Entity Recognition (NER) was defined over 20 years ago at the Message Understanding Conference \cite{muc6,Sundheim:1995:ORM:1072399.1072402}. Nowadays, there are a lot of solutions capable of a very high accuracy even on very hard and multi-domain data sets \cite{survey-C18-1182, LiAixin2018}.

Many of these solutions benefit from large available data sets or from recent developments in deep neural networks. However, in order to progress further with this last mile, we need a better understanding of the sources of errors in NER problem; as it is stated that \textit{''The first step to address any problem is to understand it''}. We performed a detailed analysis of errors on the popular CoNLL 2003 data set \cite{W03-0419}. 

Of course, different models make different mistakes. Here, we have focused on models that constitute a kind of breakthrough in the NER domain. These models are: Stanford NER \cite{stanfordNER}, the model made by the NLP team from Carnegie Mellon University (CMU) \cite{lample}, ELMO \cite{elmo}, FLAIR \cite{flair} and BERT-Base \cite{BertLM}.
In the Stanford model, Conditional Random Fields (CRF) with manually created features were tackled. Lample and the team (at CMU) used an LSTM deep neural network with an output with CRF for the first time. ELMO and FLAIR are new language modeling techniques as an encoder, and LSTM with a CRF layer as an output decoder. A team from Google used a fine-tuning approach with the BERT model in a NER problem for the first time, based on a Bi-diREctional Transformer language model (LM). 

We analyzed the data set from a linguistic point of view  in order to understand problems at a deeper level. As far as we know only a few studies analyse in details errors for NER problems~\cite{niklaus-etal-2018-survey, abudukelimu-etal-2018-error, Ichihara2015ErrorAO}. They mainly explore a range of name entities (boundaries in a text) and the precision and popular metrics of a class prediction (precision, recall, F1). We found the following discussions valuable:
\begin{itemize}
  \item \cite{abudukelimu-etal-2018-error} on annotation and extraction of Named Entities,
  \item \cite{brasoveanu-etal-2018-framing} on an analysis of errors in Named Entity Linking systems,
  \item \cite{Manning:2011:PT9:1964799.1964816} on linguistic limitations in building a perfect Part-of-Speech Tagger.
\end{itemize}

We took a different approach. First, our team of data scientists and linguists defined 4 major and 11 minor categories of types of problems typical for NLP (see Tab.~\ref{tab:categories}). Next, we acquired all erroneous samples (containing errors in model outputs) and we assigned them to the newly defined categories. Finally, we characterized the incorrect output of the models with regard to gold standard annotations and following our team's consensus. 

Accordingly, our overall contribution is a conceptualization and classification of the roots of problems with NER models as well as their characterization. Moreover, we have prepared new diagnostic sets for some of our categories so that other researchers can check the weakest points of their NER models.

In the following sections, we introduce our approach regarding the re-annotation process and model evaluation (section~\ref{sec:method}); we also show and discuss the results (section~\ref{sec:results}). Finally, we conclude our paper with a discussion (section~\ref{sec:improvements})  and draw conclusions (section~\ref{sec:conclusions}).

\section{Method}
\label{sec:method}
We commenced our research by reproducing the selected models for the CoNLL 2003 data set\footnote{ The details of the model parameters are described in our supplementary materials.}. Then, we analysed the erroneous samples, sentences from the test set. It is worth mentioning that we analysed the most common types of named entities, i.e. PER - names of persons, LOC - location names, ORG - organization names. Having  several times reviewed the model results and the error-prone data set, we defined the linguistic categories that are the most probable sources of model mistakes. As a result, we were able to annotate the samples with these categories; we then analysed the results and found a few possible improvements.

\subsection{Models description}

A brief history of the key developments of NER models for the CoNLL data is listed in Table~\ref{tab:old-model-results}. In our analysis, we chose 5 models (bold in the table) that make up significant progress.

Stanford NER CRF was the first industry-wide library to recognize NERs~\cite{stanfordNER}. The LSTM layer put forward by Lample from Carnegie Mellon University (CMU) was the first deep learning architecture with a CRF output layer~\cite{lample}. The following: a token-based language model (LM) with bi-LSTM with CRF (ELMO)~\cite{elmo}, a character-based LM with the same output (FLAIR)~\cite{flair} and a bi-directional language model based on an encoder block from the transformer architecture (BERT) with a fine tune classification output layer~\cite{BertLM} are very important techniques; and that not only in the domain of NER.

\begin{table}[t!]
\centering 
\begin{tabular}{p{5.5cm}p{1cm}}
  \hline
  Model & F1 \\
  \hline
 Ensemble of HMM, TBL, MaxEnt, RRM~\cite{Florian} & 88.76 \\
 \hline
 Semi-supervised learning~\cite{Ando:2005} & 89.31\\
 \hline
 \textbf{Stanford} CRF~\cite{stanfordNER} & 87.94\\
 \hline
 Neural network~\cite{Collobert:2011} & 89.59\\
 \hline
 CRF \& lexicon embeddings~\cite{W14-1609} & 90.90 \\
 \hline
 \textbf{CMU} LSTM-CRF~\cite{lample} & 90.94\\
 \hline
 Bi-LSTM-CNNs-CRF~\cite{P16-1101} & 91.21\\
 \hline
 \textbf{ELMO}: Token based LM Bi-LSTM-CRF~\cite{elmo} & 92.22\\
 \hline
 \textbf{BERT-base}: Fine tune Bi-Transformer LM with BPE token encoding~\cite{BertLM} & 92.4 (*) \\
 
 \hline
 CVT: Cross-view training with Bi-LSTM-CRF~\cite{D18-1217} & 92.61\\
 \hline
 BERT-large: Fine tune Bi-Transformer LM with BPE token encoding \cite{BertLM} & 92.8 (*) \\
 \hline
 \textbf{FLAIR}: Char based LM + Glove with Bi-LSTM-CRF~\cite{flair} & 93.09 (**)\\
 \hline
 Fine tune Bi-Transformer LM with CNN token encoding \cite{cloze-driven_self-attention_networks} & 93.5 \\
  \hline
\end{tabular}
\caption{Results reported in authors' publications about NER models on the original CoNLL 2003 test set. (*) There is no script for replicating these results and also hyper-parameters were not given. See a discussion at~\cite{git-bert} (**) This result was not achieved with the current version of the library. See a discussion at~\cite{git-flair} and the reported results at~\cite{flair2}}
\label{tab:old-model-results}
\end{table}

\subsection{Linguistic categories}
\label{sec:lc}

From a human perspective, the task of NER involves several sources of knowledge: the situation in which the utterance was made, the context of other texts and utterances in the particular domain, the structure of the sentence, the meaning of the sentence, and general knowledge about the world.

While designing categories for annotation, we tried to define these layers of NEs understanding; however, some of them are particularly problematic. For example, there is a problem with a distinction between the meaning (of lexical items and of a whole sentence) and general knowledge. Since there is an enormous and relentless linguistic and philosophical debate on this topic \cite{stanfordEnc}, we decided not to delimit these categories and not to distinguish them. Therefore, they have been labeled together as 'sentence level context' (SL-C).

Consequently, we ended up with a set of categories for annotating the items (sentences) from our data set, which are presented in Table~\ref{tab:categories} as well as described briefly in the following sections and more precisely in the supplementary materials. We have also added more examples for each category in this material. 

\begin{table}[ht]
\centering
\begin{tabular}{p{1.3cm}p{0.0cm}p{5.3cm}}
  \hline
  \hline
    shortcut && linguistic property \\ 
  \hline
   \hline
    DE- && Data set Errors \\ 
  \hline    
    DE-A && Annotation errors\\
    DE-WT && Word Typos\\
    DE-BS && Word/Sentence Bad Segmentation\\
  \hline
  \hline
    SL- && Sentence Level dependency \\ 
  \hline
    SL-S && Sentence Level Structure\\
    SL-C && Sentence Level Context\\
  \hline
  \hline
    DL- && Document Level dependency \\ 
  \hline
    DL-CR && Document Co-Reference\\
    DL-S && Document Structure\\
    DL-C && Document Context\\
  \hline
  \hline
    G- && General properties \\ 
  \hline
    G-A && General Ambiguity \\
    G-HC && General Hard Case \\
    G-I && General Inconsistency \\
   \hline
\end{tabular}
\caption{Linguistic categories prepared for our annotation procedure.}
\label{tab:categories}
\end{table}

\textbf{DE-A: Annotation errors} are obvious errors in the preliminary annotations (the gold standard in the CoNLL test data set). For example: in the sentence \emph{''SOCCER - JAPAN GET LUCKY WIN, CHINA IN SURPRISE DEFEAT''} as a gold standard annotation \emph{''CHINA''} is assigned a person type; it should, however, be defined as a location so as to be consistent with the other sentence annotations.

\textbf{DE-WT: Word typos} are simple typos in any word in a sample sentence, for exmple: \emph{''Pollish''} instead of \emph{''Polish''}.

\textbf{DE-BS: Word-sentence bad segmentation}. We annotated this case if a few words, joined together with a hyphen or separated by a space, were incorrectly divided into tokens (e.g. \emph{''India-South''}), or where a sentence was erroneously divided inside a boundary of a named entity, which prevented its correct interpretation. For example: in the data set there is a sentence divided into two parts: \emph{''Results of National Hockey''} and \emph{''League''}.

\textbf{SL-S: Sentence level structure} dependency occurs when there is a special construction within a sentence (a syntactic linguistic property) that is a strong premise for defining an entity. In the studied material, we distinguished two such constructions: brackets and bullets. The error receives the SL-S annotation, when the system should have been able to recognize a syntactic linguistic property that leads to correct NER tagging but failed to do so and made a NER mistake. For example: one of the analysed NER systems did recognize all locations except \emph{''Philippines''} in the following enumerating sentence: \emph{''ASEAN groups Brunei, Indonesia, Malaysia, the Philippines, Singapore, Thailand and Vietnam.''}.

\textbf{SL-C: Sentence level context} cases are those in which one is able to define an appropriate category of NE based only on the sentence context. For example: one of NER systems has a problem with recognizing the organization \emph{''Office of Fair Trading''} in the sentence: \emph{''Lang said he supported conditions proposed by Britain's Office of Fair Trading, which was asked to examine the case last month.''}.

\textbf{DL-CR: Document level co-reference} category was annotated if there was a reference within a sentence to an object that was also referred to in another sentence in the same document. For example: evaluating the \emph{''Zywiec''} named entity in the sentence \emph{''Van Boxmeer said Zywiec had its eye on Okocim ...''}, it has to be considered that there is another sentence in the same document in the data set that explains the organization name, which is: \emph{''Polish brewer Zywiec's 1996 profit...''}.

\textbf{DL-S: Document level structure} cases are those in which the structure of a document plays an important role, i.e. the occurrence of objects in the table (for example the headings determine the scope of an entity itself and its category). For example: look at the following three sentences, which obviously compose a table: \emph{''Port Loading Waiting''}; \emph{''Vancouver 5 7''}, \emph{''Prince Rupert 1 3''}. One of our NER systems had a problem with recognizing each localisation inside the table; however, the system recognized the header as a named entity.

\textbf{DL-C: Document level context} is a type of a linguistic category in which the entire context of a document (containing an annotated sentence) is needed in order to determine a category of an analysed entity, and in which none of the sentence level linguistic categories has been assigned (neither SL-S and SL-C). 

\textbf{G-A: General ambiguity} are those situations in which an entity occurs in a different sense from that in which this word (entity) is used in its most common understanding and usage. For example: the common word \emph{'pace'} may as well be occur to be a surname, as in the following sentence: \emph{''Pace, a junior, helped Ohio State...''}.

\textbf{G-HC: General hard cases} are cases occurring for the first time in a set in a given sub-type, and which can be interpreted in two different ways. For example: \emph{''Real Madrid's Balkan strike force...''} where the word \emph{'Balkan'} can be a localisation or an adjective.

\textbf{G-I: General inconsistency} are cases of inconsistencies in the annotation (in the test set itself as well as between the training and test sets). For example in the sentence: \emph{''... Finance Minister Eduardo Aninat said.''}, the word \emph{'Finance'} is annotated as an organisation but in the whole data set the names of ministries are not annotated in the context of the role of a person.

\subsection{Annotation procedure}
All those entities that had been incorrectly recognized by any of  the tested models—false positives, false negatives and wrongly tagged entities were annotated in our research by two teams. Each team consisted of a linguist and a data scientist. We did not analyse errors with the MISC entity type, but the person, localisation and organisation names. The MISC type comprises a variety of NERs that are not of other types. Its definition is rather vague and it is hard to conceptualize what it actually means, e.g. if whether it comprises events or proper names, or even adjectives. 

The annotation process was performed in four steps:
\begin{enumerate}
    \item a set of linguistic annotation categories was established, see the previous section~\ref{sec:lc};
    \item the data set was split into two equal parts: one part for each team; all entities were annotated twice, by a linguist and by a data scientist, each working independently;
    \item the annotations were compared and all inconsistencies were solved within each team; 
    \item two teams checked the consistency of the other team's annotations; all borderline and dubious cases were discussed by all team members and reconciled.
\end{enumerate}

The inter-annotator agreement statistics and Kappa are presented in Table~\ref{tab:annotators-agreements}. A few categories were very difficult to conceptualize, so it took more time to solve these inconsistencies. In these inconsistent cases, two annotators (a linguist and a data scientist) thoroughly discussed each example.

Not all categories (see Table~\ref{tab:categories}) were annotated by the whole team. Those easy to annotate, as the categories regarding simple errors (i.e. DE-A, DE-WT, DE-BS), were done by one person and then just checked by another. 

The general inconsistencies category (G-I) were done semi-automatically and then checked. The semi-automatic procedure was as follows: first finding similarly named entities in the training and test sets and then looking at their labels. By 'similarly named entities' we mean, e.g. a division of an organization having a geographical location in its name (''Pacific Division''), or a designation of a person from any country (''Czech ambassador'').

Additionally, a document level context (DL-C) category was derived from the rule of not being present in any sentence level category (i.e. SL-C or SL-S).

\begin{table}[ht]
\centering 
\begin{tabular}{p{2.4cm}p{2.4cm}p{1.5cm}}
  \hline
  annotated class & agreement [\%] & Kappa \\
  \hline
  SL-S & 94.99 & 0.572 \\
  SL-C & 69.64 & 0.389 \\
  \hline
  DL-CR & 78.00 & 0.554 \\
  DL-S & 81.44 & 0.536 \\
  \hline
  G-A & 68.96 & 0.252 \\
  G-HC & 74.46 & 0.340 \\
  \hline
\end{tabular}
\caption{Inter-annotator statistics (agreement and Kappa) at the very first stage of the annotation procedure, before discussing each controversial example and the super-annotation stage. The statistics are calculated for those categories that were annotated by human annotators.}
\label{tab:annotators-agreements}
\end{table}

\subsection{Our diagnostic procedure}
\label{sec:diagnostic-ds}

The next step, after the analysis of linguistic categories of errors, was to create additional diagnostic sets. The goal of this approach was to find, or create, more examples that reflect the most challenging linguistic properties; these can be sentence and document level dependencies and can also include a few ambiguous examples. These ambiguities are for instance names that contain words in common usage. We selected 65 examples from Wikipedia articles per two groups of linguistic problems: sentence-level and document-level contexts.\footnote{Our prepared diagnostic data sets are available at \url{https://github.com/applicaai/ner-resources}} 

The first diagnostic set comprises sentences in which the properties of a language, general knowledge or a sentence structure are sufficient to identify a NE class. We use this Template Sentences~(TS) to check whether a model will have the same quality after changing words, i.e. a name of an entity. For each sentence we prepared at least 2 extra entities with different lengths of words which are well suited to the context. For example in a sentence: \emph{''Atl\'{e}tico's best years coincided with dominant Real Madrid teams.''}, the football team \emph{''Atl\'{e}tico''} can be replaced with \emph{''Deportivo La Coru\~na''}.

The second batch of documents  was a group of sentences in which a sentence context is not sufficient to designate a NE, so we need to know more about the particular NE, e.g. we need to look for its co-references in the document, or we require more context, e.g. a whole table of sports results, not only one row. (This particular case often occurs in the CoNLL 2003 set when referring to sports results.) We called this data set Document Context Sentences~(DCS). In this data set we annotated NEs and their co-references that are also NEs. An example of such a sentence and its context is as follows: \emph{''In 2003, Loyola Academy (X, ORG) opened a new 60-acre campus ... The property, once part of the decommissioned	NAS	Glenview, was purchased by Loyola (X,ORG) in 2001.''} The~second occurrence of the \emph{''Loyola''} name is difficult to recognize as an~organization without its~first occurrence, i.e. \emph{''Loyola Academy''}. 

The~other type of~a~diagnostic set is fairly simple. It is generated from random words and letters that are capitalized or not. Its purpose is just to check if a~model over-fits a~particular data set (in~our case, the~CoNLL 2003 set). A scrutinized model should not return any entities on those Random Sentences~(RS). We generated 2 thousands of these pseudo-sentences.

\section{Results}
\label{sec:results}

\subsection{Annotation quality}
In~Table~\ref{tab:prec-model-results} we gathered our model's results for the standard CoNLL 2003 test set and the same set after the re-annotation and correction of annotation errors. We replaced only those annotations (gold standard) which we (all team members) were sure of. Those sentences in which the class of an~entity occurrence was ambiguous were not corrected. This shows that the models are better than we thought they were, and~so we corrected only the test set and left the inconsistencies.\footnote{A small part of the data set of~annotation corrections and also the debatable cases will be available at~our github -- \url{https://github.com/applicaai/ner-resources}. We decided not to open the whole data set, because it is the test set and the tuning models on this set would lead to unfair results. On the other hand, we could not perform the analysis on a validation set because it is rather poor with respect to different kinds of linguistic properties.}.

\begin{table}[ht]
\centering
\begin{tabular}{p{1.35cm}p{0.75cm}p{0.75cm}p{0.75cm}p{0.75cm}p{0.75cm}}
  \hline
   & Stan-ford & CMU & ELMO & FLAIR & BERT \\
  \hline
  ALL-O & 88.13 & 89.78 & 92.39 & 92.83 & 91.62 \\
  ALL-C & 88.73 & 90.39 & 93.21 & \textbf{93.79} & 92.33 \\ 
  \hline
  PER-O & 93.31 & 95.74 & 97.07 & 97.49 & 96.14 \\
  PER-C & 93.94 & 96.49 & 97.81 & \textbf{98.08} & 96.88  \\
  \hline
  ORG-O & 84.23 & 86.90 & 90.68 & 91.34 & 90.61  \\
  ORG-C & 84.89 & 87.53 & 91.61 & \textbf{92.64} & 91.44  \\
  \hline
  LOC-O & 90.83 & 92.02 & 93.87 & 94.01 & 92.85  \\
  LOC-C & 91.58 & 92.62 & \textbf{94.92} & 94.72 & 93.59  \\
  \hline
  MISC-O & 79.10 & 77.31 & 82.31 & 82.89 & 80.81  \\
  MISC-C & 79.37 & 77.58 & 82.47 & \textbf{84.40} & 81.10  \\
\end{tabular}
\caption{Results for selected models on the original (designated as ending '...-O') and re-annotated / corrected ('...-C') CoNLL 2003 test set concerning NE classes (ALL comprise PER, ORG, LOC, MISC). The given metric is a multilabel-F1 score (percentages).} 
\label{tab:prec-model-results}
\end{table}

\subsection{Linguistic categories statistics}

\begin{figure}[h!t]
\includegraphics[width=7.5cm]{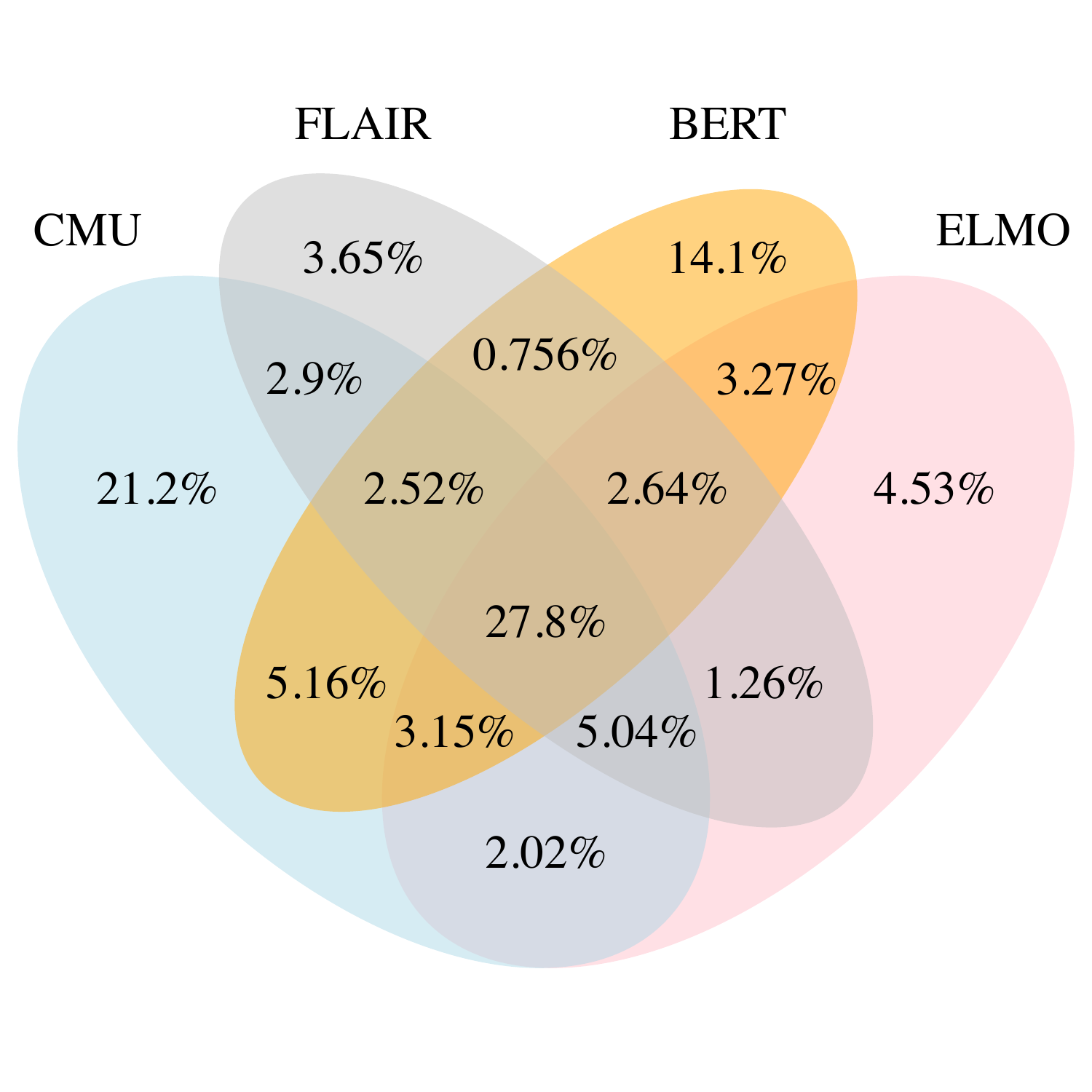}
\caption{Venn diagram for errors in the CMU, FLAIR, BERT, ELMO models. The four models generate 794 errors and 221 are common to all of them. The Stanford model as the most error-prone is here not referred to.}
\label{fig:mistakes-venn}
\end{figure}

\begin{figure}[ht]
\includegraphics[width=8cm]{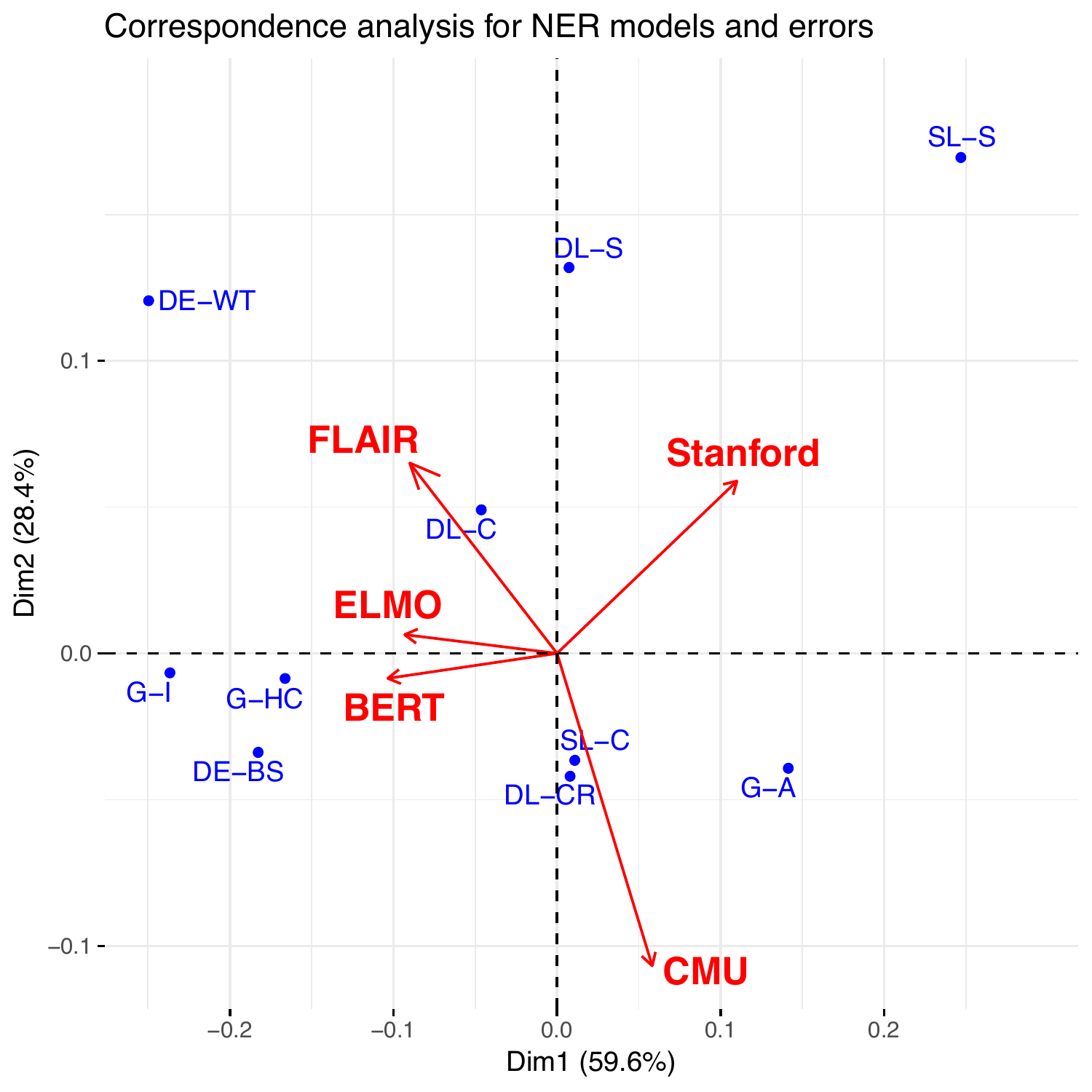}
\caption{Correspondence analysis for the models' errors. ELMO, FLAIR and BERT are more affected by G-HC and G-I, FLAIR is also reduced with DL-C and DE-WT. See Table~\ref{tab:errors} for more details and Table~\ref{tab:categories} for names of categories.}
\label{fig:mistakes-coresp}
\end{figure}

In the~CoNLL 2003 test set, we chose as samples words and sentences in which at~least one model made a mistake. The set of errors comprises 1101 named entities. The results of each model on this set in terms of our linguistic categories are presented in~Fig.~\ref{fig:mistakes-venn},  Fig.~\ref{fig:mistakes-coresp} and in~Table~\ref{tab:errors}. 

Most mistakes were made by the Stanford and CMU models, 703 and 554 respectively. ELMO, FLAIR and BERT, which use contextualised language models, performed much better. These embedded features help the models to understand words in their context and thus resolve most problems with ambiguities.

The CMU model has most problems with sentence level context and ambiguity. This is probably due to the fact that this model uses non-contextualized embedded features (Fig.~\ref{fig:mistakes-coresp}). The Stanford model fares the worst in terms of structured data (almost twice as many errors as the other models), which means that it is not good at defining an entity type within a very limited context (Tab.~\ref{tab:errors}). The Stanford model's hand-crafted features do not store information about the probabilities of words which could represent a specific entity type. It generates much more errors than the other models.

\begin{table}[ht]
\centering
\begin{tabular}{p{1.3cm}p{0.8cm}p{0.8cm}p{0.8cm}p{0.8cm}p{0.8cm}}
  \hline
 & Stan-ford & CMU & ELMO & FLAIR & BERT \\ 
  \hline
  DE-WT & 10 & 6 & 9 & 8 & 10 \\ 
  DE-BS & 38 & 39 & 33 & 33 & 40 \\ 
  SL-S & 46 & 21 & 13 & 16 & 11 \\ 
  SL-C & 448 & 378 & 250 & 223 & 300 \\ 
  DL-CR & 372 & 316 & 198 & 184 & 263 \\ 
  DL-S & 202 & 107 & 97 & 100 & 117 \\ 
  DL-C & 247 & 175 & 144 & 146 & 170 \\
  G-A & 219 & 183 & 98 & 101 & 94\\ 
  G-HC & 72 & 68 & 65 & 59 & 65 \\ 
  G-I & 19 & 20 & 21 & 20 & 20 \\
  \hline
  Errors & 703 & 554 & 395 & 370 & 472 \\
  \hline
  Unique &  &  &  &  & \\
  errors & 235 & 93 & 23 & 12 & 79 \\
  \hline
\end{tabular}
\caption{Number of errors for a particular model and a particular class of errors. The~total number of annotated errors is 1101.}
\label{tab:errors}
\end{table}

\begin{figure}[h!t]
\includegraphics[width=8cm]{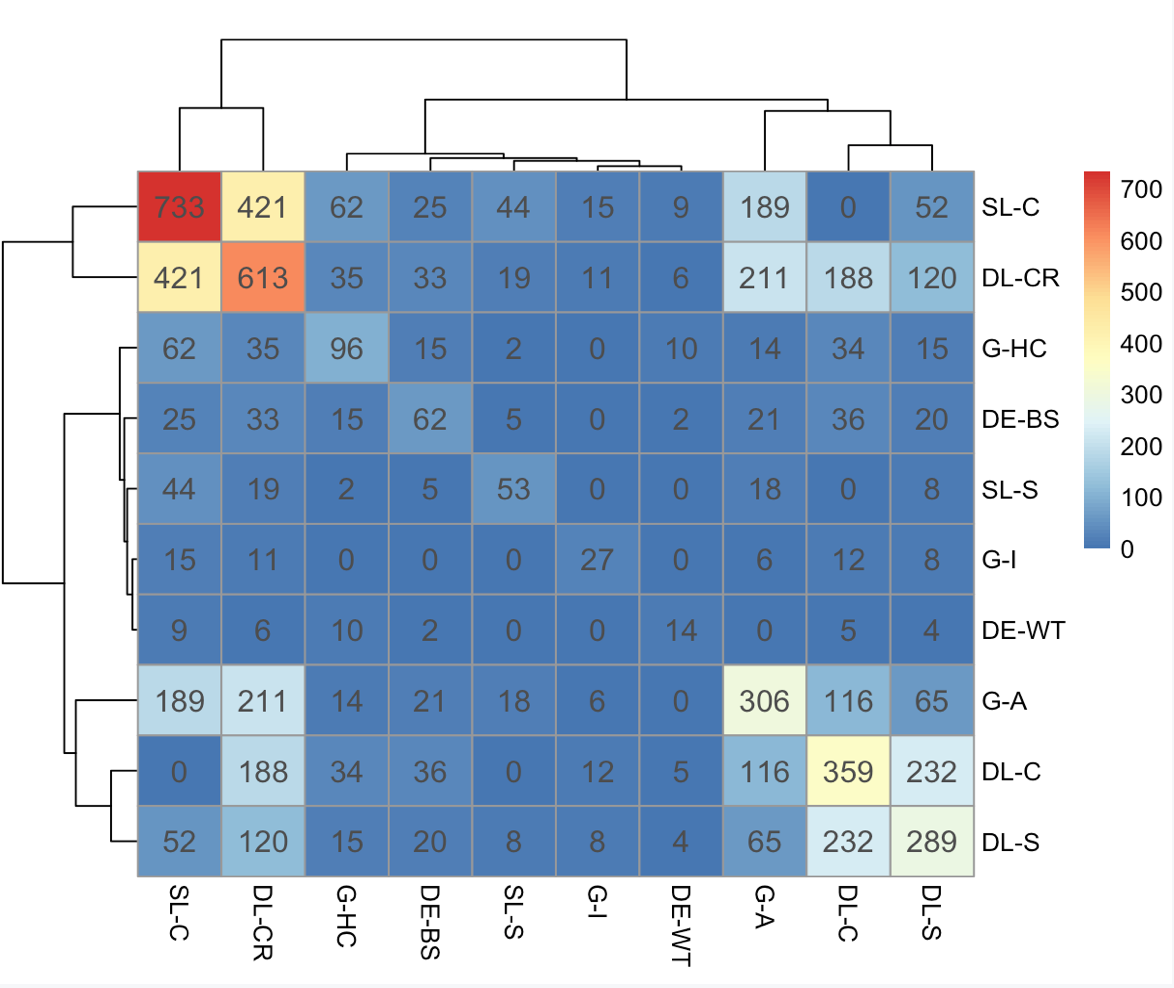}
\caption{Heatmap for errors from the five considered models. 197 errors are common to all the models. In this figure we can see which linguistic categories tend to occur together.}
\label{fig:mistakes-heatmap}
\end{figure}

\begin{figure}[h!t]
\includegraphics[width=8cm]{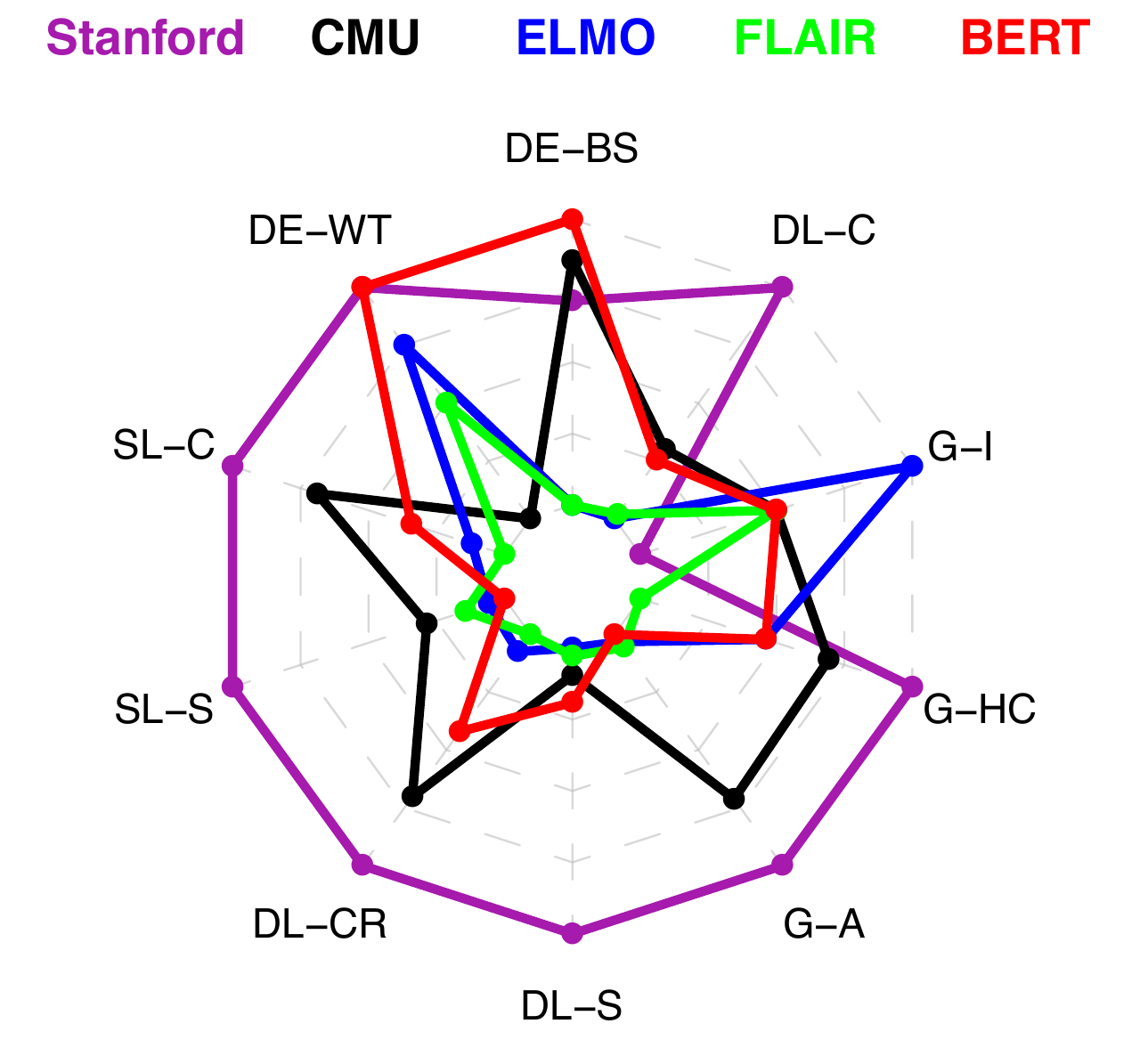}
\caption{Radar plot with the strong and weak sides of NER models. A radius corresponds to a number of errors in a given linguistic category, the smaller the better. See Table \ref{tab:errors} for more details.}
\label{fig:radar}
\end{figure}

Modern techniques using contextualized language models like ELMO, FLAIR and BERT reduced a number of mistakes in SL-C category by more than 50\% in comparison to the Stanford model. But they are unable to fix most errors in general problems related to inconsistency (G-I), general hard cases (G-HC) or word typos (DE-WT). See Figure \ref{fig:radar} for more details.

Nevertheless, there are still a lot of common problems (27.8\%). In~common errors (Fig.~\ref{fig:mistakes-heatmap}), SL-C (sentence level context) and DL-CR (document level co-reference) co-occur the~most often. Thus, if a~model also takes into account the~context of a~whole document, it can be of~great benefit. Considering a document structure (DL-S) in modeling is also very important. This also can help to resolve a lot of ambiguity issues (G-A). Here is an example of such a situation: \emph{''Pace outdistanced three senior finalists...''}, \emph{''Pace''} is a person's surname, but one is able to find it out only when analysing the whole document and finding references to it in other sentences that directly point to the class of the named entity. 

We must be aware of the fact that some problems cannot be resolved with this data set, not even in general. Those problems have roots in two main areas: data set annotation (word typos, bad segmentation, inconsistencies) and a complicated structure of a language. Generally in most languages it is easier to say what entity represents a real word instance than to define an exact entity type (especially when we use a metonymic sense of a word), e.g. 'Japan' can be a name of a country or of a sports team.

\subsection{Diagnostic data sets}

Looking at the models' results in our diagnostic data sets (Tab.~\ref{tab:diagnostic-datasets-results}), the first and most important observation is that we achieved significantly lower results than originally on the CoNLL 2003 test set\footnote{We add statistics and a few examples from our diagnostic data sets in the supplementary materials.}. The reason for this is that diagnostic examples were selected for a broader range of topics (not only politics or sports). In~particular, document context sentences (DCS) contain 364 unique entities of which only 47 appeared in an exact word form in the training data, and only 42 of them have the same entity type (organization, location or person) - the same type as in the CoNLL 2003 training set. Additionally, those sentences are also difficult due to their linguistic properties (for some entities you must analyse a whole article to properly distinguish  their type). 

As far as the results of the diagnostic sets are concerned, we observed much better results for solutions using embeddings generated by the language models. It seems that by using ELMO embeddings we can outperform the FLAIR and BERT-Base models in case of sentences about general topics, in which the context of a whole sentence is more important than properties of words composing entities. 

Moreover, when we tested all the models on random sentences (RS), this was not so good as we might have expected. All the models are very sensitive to words starting with or consisting of capital letters. Results from this diagnostic set could help to choose a model that must work properly on documents which were produced by the OCR engine with their many mistakes and misspellings. 

Another interesting idea is to train or just test a model on some template sentences (TS). With such a data set we can test a model's ability to detect proper boundaries of an entity. We can do it by replacing a template entity with another one consisting of a different number of words. We could also adjust our models to a particular domain, e.g.to change entities with a PERSON type in an original data set to be more globally diversified, if we have to extract person names from the whole world (Asian or Russian names).  


\begin{table}[ht]
\centering
\begin{tabular}{p{1.6cm}p{0.7cm}p{0.7cm}p{0.8cm}p{0.8cm}p{0.8cm}}
  \hline
   & Stan-ford & CMU & ELMO & FLAIR & BERT \\
  \hline
  DCS (F1)  & 45.37 & 61.86 & \textbf{76.36} & 71.89 & 68.90 \\
  DCS (P)  & 43.66 & 58.07 & 73.11 & 69.35 & 59.06 \\
  DCS (R)  & 47.21 & 66.17 & 79.92 & 74.63 & 82.66 \\
  \hline
  TS-O (F1)  & 68.96 & 79.66 & \textbf{89.45} & 88.51 & 83.47 \\
  TS-O (P)  & 76.92 & 78.33 & 85.48 & 85.25 & 75.18 \\
  TS-O (R)  & 62.50 & 81.03 & 93.81 & 92.04 & 93.81 \\
  \hline
  TS-R (F1)  & 63.06 & 72.86 & 85.01 & \textbf{86.63} & 79.66 \\
  TS-R (P)  & 65.47 & 70.65 & 81.45 & 83.70 & 71.60 \\
  TS-R (R)  & 60.83 & 75.21 & 88.91 & 89.77 & 89.77 \\
  \hline
  RS (No)  & 3571 & 3339 & 2096 & \textbf{1404} & 3086  \\
  \hline
\end{tabular}
\caption{Diagnostic data sets results for selected models: 'DCS' - Document Context Sentences, 'TS-O' - Template Sentences with original entities, 'TS-R' - Template Sentences with replaced entities,  'RS' - Random Sentences. F1=multilabel F1-score, P=Precision, R=Recall, No=number of returned entities (lower is better). In the RS data set there are 2000 strings pretending to be sentences.}
\label{tab:diagnostic-datasets-results}
\end{table}

\section{Discussion}
\label{sec:improvements}

On the basis of our research, we can draw a number of conclusions that are not often addressed to in publications about new neural models, their achievements and architecture. The scope of any assessment of new methods and models should be broadened to the understanding of their mistakes and the reasons why these models perform well or poorly in concrete examples, contexts and word meanings. These issues are particularly important in text data sets, in which semantic meaning and linguistic syntax are very complex.

In our effort to define linguistic categories for problematic Named Entities and their statistics in the CoNLL 2003 test set, we were able to draw a few additional conclusions regarding data annotation and augmentation processes. Moreover, our categories are similar to the taxonomy defined in publication about errors analysis for Uyghur Named Tagger \cite{abudukelimu-etal-2018-error}. 

\subsection{The annotation process}

The annotation process is a very tedious and exhaustive task for a person involved. Errors in data sets are expected but what must be checked is their impact on generalizing a model, e.g. one can create entities in places where they do not occur and check the model's stability. There are some useful applications for detecting annotation errors  \cite{Snorkel}, \cite{gralinski-etal-2019-geval} and \cite{wisniewski-2018-errator} but they are not used very often. Obviously, an appropriate and exhaustive documentation for the data set creation and annotation process is crucial. All annotated entity types should be described in details and examples of border cases should be given. 
In our analysis of the CoNLL 2003 data set we did not find any documentation. We have made our own assumptions and tried to guess why some classes are annotated in a given way. However, the work was hard and required many discussions and extended  reviews of literature.

Secondly, there is a need for extended data sets with a broadened annotation process, similar to that of our diagnostic sets. E.g. linguists can extend their work not only just to the labelling of items (sentences), but also to indicating the scope of context that is necessary to recognise an entity, and to extending annotations for difficult cases or adding sub-types of entities. 

Our work on diagnostic data sets is an attempt to extend an annotation process by focusing only on specific use cases which are less represented in the original data set.  

\subsection{Extended context}

A new model training process itself should consist of more augmentation of the data set. Currently, there is some work being done on this topic, e.g. a semi-supervised context change with cutting the neighbourhood around NEs using a sliding window~\cite{D18-1217}. Other techniques could be a random change of the first letter (or whole words) of NEs so that the model would not be so vulnerable to capitalized letters in names or small changes in sentences (e.g. adding or removing a dot at the end of a sentence). 

Furthermore, a sentence itself is not always sufficient to recognise a class of a NE. In these cases, in both training and test data sets, there should be more samples where there are indications of co-references that are important to recognise particular NEs. Then, the input of a model should comprise a sentence and embedded features (or any representation) of co-references or their contexts. E.g. \emph{''Little was banned. Peter Little took part in the last match with Welsh team.''} - in the first sentence, we are are not sure if it is a NE. Then \emph{''Peter Little''} indicates the proper NE type. An example of a model and data processing pipeline (i.e. memory of embeddings) that takes into consideration the same names in different sentences is to be found in~\cite{flair2} and~\cite{zhang2018global}.

Another important improvement is adding information about document layout or the structure of a text, e.g. a table, its rows and columns, and headings. In CoNLL 2003, there are many sports news, stock exchange reports or timetables where the structure of a text helps to understand its context, and thus to better recognise its NEs. Such a solution for another domain—invoice information extraction—is elaborated on by~\cite{chargrid} or~\cite{DBLP:journals/corr/abs-1903-11279}. The solutions mentioned here combine character information with document image information in one architecture of a neural network.

The CoNLL 2003 test set is certainly too small to test the generalisation and stability of a model. Faced with this issue, we must find new techniques to prevent over-fitting. For instance, we could check a model's resistance to examples prepared in our diagnostics data sets, e.g. after changing a NE in a template sentence, the model should find the entity in the same place. We could also prepare small modifications to our original sentences, e.g. add or remove a dot at the end of an example and compare results (similarly to adversarial methods). 

\section{Concluding remarks}
\label{sec:conclusions}
Mistakes are not all created equal. A comparison of models based on scores like F1 is rather simplistic. In this paper we defined 4 major and 11 minor linguistic categories of errors for NER problems. For the CoNLL 2003 data set and five important ML models (Stanford, CMU, ELMO, FLAIR, BERT-base) we re-annotated all errors with respect to the newly proposed ontology. 

The presented analysis helps better understand a source of problems in recent models and also to better understand why some models are more reliable on one data set but less not on another. 

\section*{Acknowledgements}

TSt, AWr, AWó, DZi would like to thank for the financial support that their project conducted by Applica.ai has received from the European Regional Development Fund (POIR.01.01.01-00- 0144/17-00). PBi was financially supported by European Regional Development Fund POIR.01.01.01-00-0328/17.

Our research was also partially supported as a part of the RENOIR Project by the European Union’s Horizon 2020 Research and Innovation Programme under the Marie Skłodowska-Curie grant agreement No 691152 and by the Ministry of Science and Higher Education (Poland), grant Nos. W34/H2020/2016.

\bibliography{emnlp-ijcnlp-2019}
\bibliographystyle{acl_natbib}

\end{document}